\documentclass{article} %
\usepackage{iclr2027_conference,times}

\usepackage{amsmath,amsfonts,bm}

\def\eqref#1{equation~\ref{#1}}

\def\1{\bm{1}}

\DeclareMathAlphabet{\mathsfit}{\encodingdefault}{\sfdefault}{m}{sl}
\SetMathAlphabet{\mathsfit}{bold}{\encodingdefault}{\sfdefault}{bx}{n}

\usepackage{hyperref}
\usepackage{cleveref}
\usepackage{url}
\usepackage{graphicx}
\usepackage{enumitem}
\usepackage{wrapfig}
\usepackage{xcolor}
\usepackage{comment}
\usepackage{booktabs}
\usepackage[normalem]{ulem}
\usepackage{amsthm}
\newtheorem{assumption}{Assumption}[section] 
\newtheorem{remark}{Remark}[section] 
\newtheorem{proposition}{Proposition}[section] 
\newtheorem{corollary}{Corollary}[section]
\usepackage{tikz}
\usetikzlibrary{arrows.meta,positioning,shapes.geometric,calc}
\definecolor{ink}{HTML}{1B2430}\definecolor{mut}{HTML}{5B6673}\definecolor{rule}{HTML}{9CA3AF}
\definecolor{rb}{HTML}{2A6FDB}\definecolor{rbl}{HTML}{EAF2FD}
\definecolor{am}{HTML}{C9720A}\definecolor{aml}{HTML}{FDF1DF}
\definecolor{gr}{HTML}{F3F5F7}\definecolor{gn}{HTML}{2E8B57}

\DeclareRobustCommand{\circled}[1]{%
  \tikz[baseline=(c.base)]{%
    \node[
      circle,
      fill=rb,
      text=white,
      font=\scriptsize\bfseries,
      inner sep=0pt,
      minimum size=1.1em
    ] (c) {#1};
  }%
}

\DeclareRobustCommand{\circled}[2][rb]{%
\tikz[baseline=(c.base)]{%
\node[circle,fill=#1,text=white,font=\scriptsize\bfseries,inner sep=0pt,minimum size=1.1em] (c) {#2};
}%
}

\title{ORACLE: Agentic AI Orchestrator Routing via Adaptive Verifier Calibration Feedback}

\newcommand{\NAME}{ORACLE }
\newcommand{\NAMEnospace}{ORACLE}

\author{
{\normalfont\mdseries
Ritik Raj\textsuperscript{1,*}, 
Souvik Kundu\textsuperscript{2,*}, 
Dheemanth Joshi\textsuperscript{3}, 
Tushar Krishna\textsuperscript{1}
}
\\[2mm]
\textsuperscript{1}Georgia Institute of Technology
\qquad
\textsuperscript{2}Intel 
\qquad
\textsuperscript{3}Texas A\&M University
\\
\textsuperscript{*}Correspondence:
\texttt{ritik.raj@gatech.edu},
\texttt{souvikk.kundu@intel.com}
}

\iclrfinalcopy

\begin{document}

\maketitle

\begin{abstract}

Modern enterprise agent deployments consist of a heterogeneous pool of large language models (LLMs) having diverse capabilities and cost. Existing model routing strategies optimize the quality-cost trade-off, while providing request-level \textit{static} decisions. More recent solutions address agentic routing as a task-level selection with a serial verifier based router feedback loop. However, their fixed verifier suitable for homogeneous workloads may not generalize to heterogeneous batches of agentic tasks (example: coding, general conversational). Additionally, due to the verifier placement in the critical path of the loop, serving quality may be affected during multiple concurrent requests routing. To mitigate these issues, we present \NAMEnospace. It is a concurrency-aware online routing mechanism that aligns adaptive routing with adaptive verification for feedback. \NAME acts as a \textit{training-free} drop-in `feedback loop' on top of any model-selection policy to first classify the task type and then dynamically assigns a task-appropriate verifier. Further, we develop a \textit{delayed feedback} strategy for concurrent requests that largely removes verifier latency from the dispatch critical path. We then present a post-routing \underline{di}spatch \underline{sc}heduler, namely DISC. DISC reserves each task's peak KV footprint at admission and dispatches to an alternate backend when the reward gain from reduced wait exceeds the reward loss from lower accuracy. Extensive evaluation on SWE-bench, $\tau^2$-bench, and Terminal-Bench~2.0 shows that \NAME improves the accuracy--cost frontier by up to 7 percentage points over state-of-the-art routing baselines, while \NAME with DISC improves program throughput by up to $1.8\times$. To facilitate reproducibility and support future development, we open-source the code-base of \NAME at \href{https://github.com/ORACLE-org/ORACLE}{https://github.com/ORACLE-org/ORACLE}.

\end{abstract}

\section{Introduction}
\label{sec:introduction}

Modern enterprise agent deployment involves a pool of large language model (LLM) choices. For example, \citet{datadog2026state} reports that over 70\% of the organizations use three or more models. In such deployments, larger models generally offer higher accuracy; however, they are costly and slow. Smaller models trade accuracy for lower cost and faster inference. Model serving APIs from different vendors may include models of similar sizes that provide different accuracy–cost trade-offs, with relative performance varying across benchmarks. Efficient and effective routing has therefore become an integral part of the agent orchestration layer, for optimized serving of agentic multi-turn ReAct \citep{yao2022react} style tasks. For example, LLMRouterBench~\citep{li2026llmrouterbench} reports that routing across heterogeneous models can reduce cost by up to 31.7\% without sacrificing accuracy.

The majority of routing works such as different variants of the semantic router~\citep{liu2026vllm, xue2026r2, tian2026rankguide} make model-selection decisions from request features without adapting to execution outcomes. This problem escalates in multi-turn agentic use cases, as the router's ability to select can have a domino effect on both cost and efficacy of the agent-harness. We thus argue that \textit{routing principle needs to learn from the execution feedback}. A recent work, ACRouter~\citep{zhou2026acrouter}, is consistent with this principle, incorporating execution-grounded feedback into subsequent decisions through a \textit{sequential} context--action--feedback (CAF) loop. However, this work faces two key challenges. First, the lack of task classification can corrupt online learning because heterogeneous tasks, such as enterprise tool use and software engineering, require different feedback mechanisms. Second, its sequential CAF loop can place verification (Sec.~\ref{sec:related}) on the critical path under concurrent request handling.

Beyond feedback adaptation, heterogeneous routing also introduces a systems challenge: \textit{load imbalance}. Routing decisions can concentrate requests on one backend (a serving instance hosting a particular model) while leaving another underutilized. The resulting increase in backend concurrency can degrade KV-cache efficiency and throughput: ThunderAgent~\citep{kang2026thunderagent} reports that, even for homogeneous models, scaling from 24 to 96 concurrent agentic tasks reduces prefix-cache hit rate from roughly 98\% to 32\% and throughput from about 220 to 140 steps/min.

\begin{figure}
        \centering
    \includegraphics[width=0.9\linewidth]{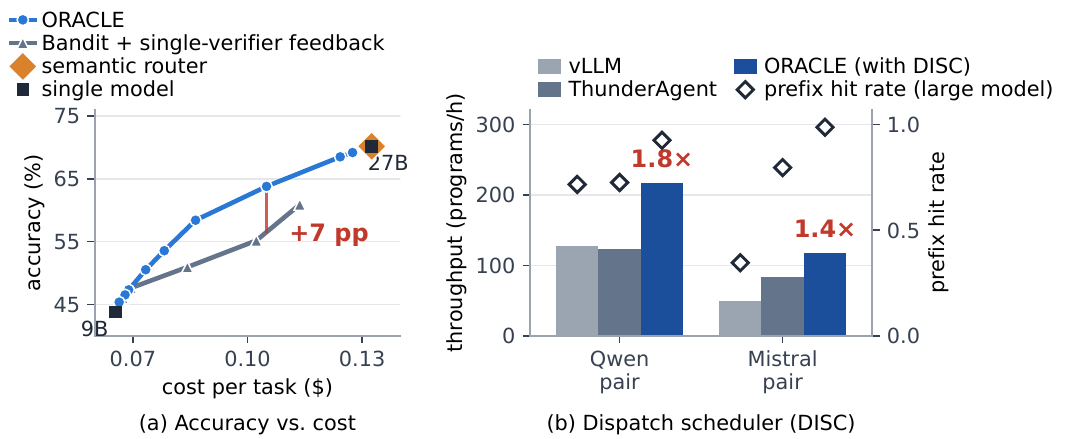}
\caption{
\NAME comparison with routing and scheduling baselines using
Qwen3.6-27B (FP8) and Qwen3.5-9B; panel (b) adds Devstral-Small-2-24B and Ministral-3-14B.
(a) Accuracy--cost trade-off on a heterogeneous stream of
SWE-bench Verified, $\tau^2$-bench, and Terminal-Bench~2.0.
\NAME uses the same bandit selector as the baseline but replaces
single-verifier feedback with adaptive verifier feedback, achieving up to
7\,pp higher accuracy at matched cost.
(b) Program-throughput comparison on concurrent SWE-bench Verified
tasks, where \NAME with DISC achieves 1.4--1.8$\times$ higher
throughput than ThunderAgent.}
\vspace{-5mm}
    \label{fig:teaser}
\end{figure}

\textbf{Our Contributions.} We present \NAMEnospace, a concurrency-aware online routing mechanism that provides a \textit{training-free}, drop-in feedback loop over existing model-selection policies. \NAME first identifies each task's type and assigns a corresponding verifier, whose outcome feedback guides subsequent routing decisions. At each task's first request, \NAME stores the selected model and verifier in a program-ID-indexed lookup table; subsequent requests reuse this binding. This abstraction decouples \NAMEnospace's feedback loop from the underlying routing policy (e.g., Agent router, complexity router) and verifiers (e.g., LLM-as-a-judge, unit tests), enabling seamless integration. 

Building on this lookup-table abstraction, \NAME supports concurrent heterogeneous tasks through two mechanisms: \textit{adaptive verification} and \textit{delayed feedback}. Adaptive verification is driven by task classification: \NAME infers each task's type from its initial request and selects a corresponding verifier for outcome feedback. With delayed feedback, \NAME dispatches new tasks using the current routing policy without waiting for prior verification. Feedback from completed tasks asynchronously updates the policy for subsequent decisions, removing verification from the dispatch critical path. 

We further present DISC, a post-route dispatch scheduler that mitigates load imbalance across heterogeneous backends. It reserves each task's expected peak KV-cache footprint in a per-backend ledger, admitting tasks only when capacity permits. At task boundaries, DISC selects an alternate backend only when the predicted reward gain from reduced admission wait exceeds the estimated reward loss from lower accuracy. \NAME includes DISC by default.

We evaluate \NAME on a heterogeneous stream of SWE-bench Verified~\citep{jimenez2024swebench},
$\tau^2$-bench~\citep{barres2025tau2}, and Terminal-Bench~2.0~\citep{merrill2026terminalbench},
and on concurrent SWE-bench Verified tasks in Fig.~\hyperref[fig:teaser]{1}.
With the same bandit selector, \NAMEnospace's adaptive verifier feedback achieves up to
\textbf{7 percentage points (pp)} higher accuracy than single-verifier
feedback at matched cost (Fig.~\hyperref[fig:teaser]{1a});
with DISC, \NAME achieves \textbf{1.4--1.8$\times$} higher program
throughput than ThunderAgent across the two evaluated model pairs
(Fig.~\hyperref[fig:teaser]{1b}).

\section{Related Works}
\label{sec:related}

\paragraph{Feedback-aware routing.}
SWE-Router~\citep{son2026sweroute} uses a partial execution trajectory to decide whether to continue with a cheaper model or restart with a stronger model. ACRouter accumulates execution feedback across coding tasks through a sequential context--action--feedback loop, using task-dependent verification signals. \NAME targets concurrent, heterogeneous task streams through a \textit{training-free} feedback loop over existing model-selection policies. Its delayed feedback strategy incorporates verified outcomes into subsequent routing decisions without blocking new task dispatch.

\paragraph{Verification of agentic tasks.}
Software-engineering tasks are evaluated through test execution~\citep{jimenez2024swebench,xia2024agentless}, while LLM-based judges assess open-ended responses~\citep{zheng2023judging} and agent-based judges inspect execution artifacts against task requirements~\citep{zhuge2025agentjudge}. JudgeBench~\citep{tan2025judgebench} examines judge reliability on difficult responses, and work on noisy verification rewards~\citep{cai2026noisyverifier} studies how verifier errors affect reinforcement learning. RewardAgent~\citep{peng2025agenticrm} selects verification agents and combines factuality and instruction-following checks with preference scores for training and inference-time response selection. \NAME applies adaptive verification to online model selection: it infers each task's type from the initial request, selects an appropriate verifier, and uses the resulting outcome feedback to update the routing policy.

\paragraph{Load-aware routing and scheduling.}
ThunderAgent~\citep{kang2026thunderagent} schedules agentic programs using their KV-cache footprints and tool-resource lifecycles. Aragog~\citep{dai2025aragog} identifies workflow configurations predicted to preserve accuracy and adapts stage-level model assignments to serving capacity. DISC, the post-route scheduler in \NAMEnospace, uses per-backend reservations of each task's expected peak KV-cache footprint to control admission and predict waiting time. Before a task starts, DISC selects an alternate backend only when the predicted reward gain from reduced wait exceeds the estimated reward loss from lower accuracy.

\section{Challenges in Routing Concurrent Agentic Workloads}
\label{sec:challenges}

We examine three challenges in feedback-aware concurrent agentic routing:
verifier mismatch, feedback handling under concurrency, and load
imbalance across model backends. Our routing experiments use a
LinUCB learner in the feedback loop of a released feedback-aware
router~\citep{zhou2026acrouter}, with replayed
outcomes for Qwen3.5-9B and Qwen3.6-27B-FP8. We also profile a
program KV-aware serving system~\citep{kang2026thunderagent} on two
GH200 GPUs with 64 concurrent SWE-bench Verified tasks.

\subsection{Verifier Mismatch Limits Online Learning}

\begin{figure}
        \centering
    \includegraphics[width=0.8\linewidth]{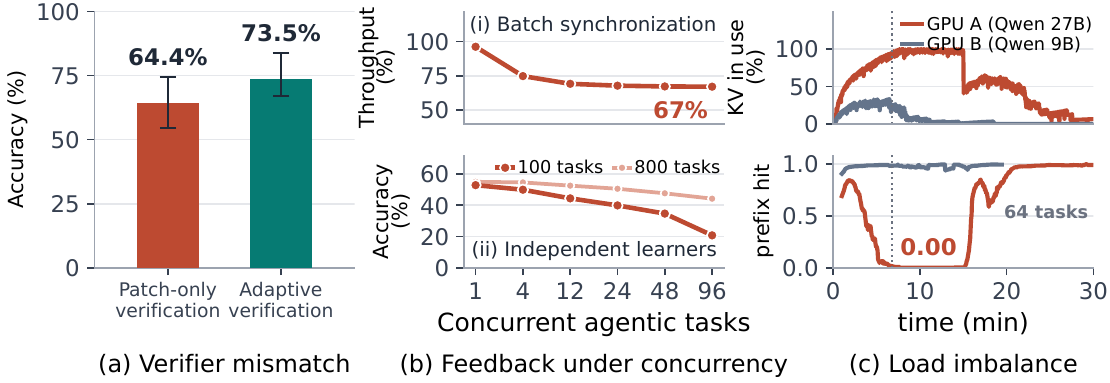}
\caption{Challenges in feedback-aware routing for concurrent agentic
workloads.
(a) Verifier mismatch: $\tau^2$-bench accuracy after eight passes over
the same mixed task stream, changing only verification feedback.
(b) Feedback under concurrency on SWE-bench Verified: batch-synchronous dispatch loses throughput and independent per-slot learners lose accuracy.
(c) Load imbalance under routing of 64 concurrent SWE-bench Verified tasks across
Qwen3.6-27B (FP8) and Qwen3.5-9B backends on two GH200 GPUs.}

    \label{fig:motivation_algo}
\end{figure}

Feedback-aware routing uses the outcomes of completed tasks to improve
model selection. Applying the same verifier across different task types,
however, can give the learner incorrect feedback. In our replay (Fig.~\hyperref[fig:motivation_algo]{2a}), a
verifier designed for software patches reports every $\tau^2$-bench task
as a failure because it produces no patch, regardless of whether the
task succeeds. We compare this patch-only setup with adaptive
verification, which selects a verifier based on the inferred task type. Both use the same learner, task orders, and mixed stream of
$\tau^2$-bench tasks and SWE-bench background tasks. We measure
$\tau^2$-bench success using its benchmark evaluator in both settings. After eight passes over the stream, adaptive verification improves
$\tau^2$-bench success by approximately 9.1 percentage points (pp)
over patch-only verification. Since only the verification feedback changes, this comparison shows
that task-appropriate verification improves routing with the same learner.

\subsection{Feedback Handling Under Concurrency}

We examine two implementations of concurrent feedback-aware routing.
\textit{Batch-synchronous dispatch} uses a shared learner but waits
for every task in a batch to finish and be verified before
dispatching the next batch. \textit{Independent per-slot learners}
allow each slot to proceed independently but divide the available
feedback: with $C$ slots and a fixed total task budget, each learner
receives approximately $1/C$ of the stream.

Fig.~\hyperref[fig:motivation_algo]{2b} evaluates both implementations
on SWE-bench Verified. We measure that batch-synchronous dispatch achieves 67\% of ideal throughput
at 96 concurrent tasks. The ideal reference starts each slot's
next task as soon as execution finishes, without waiting for
verification or other slots. With independent
learners, increasing concurrency from 1 to 96 concurrent tasks reduces task
success by approximately 32\,pp for a 100-task stream and 11\,pp
for an 800-task stream. These results motivate \textit{delayed
feedback} with a shared routing policy: new tasks proceed without
waiting for prior verification, while completed-task feedback
updates subsequent routing decisions.

\subsection{Load Imbalance Across Heterogeneous Backends}

Model-selection decisions can concentrate concurrent tasks on one
backend without accounting for its available KV-cache capacity.
The preferred backend may then become memory-constrained while
another retains spare capacity. Under high concurrency, this
imbalance can disrupt prefix reuse and reduce serving efficiency.

Fig.~\hyperref[fig:motivation_algo]{2c} shows this imbalance when
64 concurrent SWE-bench Verified tasks are routed across
Qwen3.6-27B (FP8) and Qwen3.5-9B backends without accounting
for backend load. The 27B backend approaches
full KV-cache capacity after approximately 6.8 minutes, and its
prefix-cache hit rate falls to near zero. Meanwhile, the 9B backend
reaches at most 33\% KV-cache use and maintains a prefix-cache hit
rate of roughly 98\% while active. These observations motivate
DISC's use of per-backend KV-cache reservations to guide admission
and initial task dispatch.

\section{ORACLE: A Concurrency-aware Agentic Routing System}
\label{sec:methodology}

\begin{figure}[t]
  \centering
  \resizebox{\textwidth}{!}{\begin{tikzpicture}[x=1mm, y=1mm, font=\footnotesize, >={Stealth[length=1.8mm]}, line cap=round, line join=round,
  box/.style={draw=rule, line width=.6pt, rounded corners=2pt, fill=white, text=ink, align=center, inner sep=3pt},
  sub/.style={box, font=\scriptsize, inner sep=2.5pt},
  tag/.style={circle, fill=rb, text=white, font=\scriptsize\bfseries, inner sep=0pt, minimum size=4mm},
  lab/.style={text=mut, font=\scriptsize, align=center, inner sep=1pt},
  req/.style={->, line width=.9pt, ink},
  fb/.style={->, line width=.8pt, dashed, gn},
  sig/.style={->, line width=.8pt, densely dotted, am},
]
\node[box, fill=gr, minimum width=27mm, minimum height=22mm] (harness) at (14,21) {};
\node[anchor=north, font=\footnotesize\bfseries] at ($(harness.north)+(0,-1)$) {Agent programs $\mathcal{P}$};
\node[sub, fill=white] (p1) at (7,20) {$\mathcal{P}_1$};
\node[sub, fill=white] (p2) at (14,20) {$\mathcal{P}_2$};
\node[sub, fill=white, draw=none] (p3) at (21,20) {$\cdots$};
\node[lab, anchor=south] at ($(harness.south)+(0,1)$) {requests carry the $ID$};

\node[box, draw=rb, fill=rbl, line width=.8pt, minimum width=68mm, minimum height=48mm] (router) at (68,21) {};
\node[anchor=north west, text=rb, font=\footnotesize\bfseries] at (36,44.5) {Router};
\node[anchor=north east, text=mut, font=\scriptsize] at (99.5,44.5) {$\mathcal{R}=\langle ID,\mathcal{P},\mathcal{M},\mathcal{V}\rangle$};
\node[sub, fill=gr, minimum width=26mm, align=center] (table) at (49.5,35.5) {\textbf{look-up table}\\ $ID\mapsto(m_{ID},v_{ID})$\\ \textcolor{mut}{bound once, reused}};
\node[diamond, draw=rule, fill=white, aspect=2, inner sep=.5pt, font=\scriptsize, align=center] (new) at (49.5,21) {new\\$ID$?};
\node[sub] (emb) at (61,21) {embed\\$e_{ID}$};
\node[sub, draw=rb, minimum width=29mm] (vsel) at (86,31) {\textbf{adaptive verification}\\ classify task type\\ assign verifier $v_{ID}$};
\node[sub, draw=rb, minimum width=29mm] (bandit) at (86,11) {\textbf{model selector} ($m_{ID}$)\\ agent router\\ complexity router\\ bandit router};
\node[tag] at (72.5,36) {1};

\node[box, draw=am, fill=aml, line width=.8pt, minimum width=34mm, minimum height=48mm] (sched) at (123,21) {};
\node[anchor=north west, text=am, font=\footnotesize\bfseries] at (107,44.5) {DISC scheduler};
\node[tag, fill=am] at (137.5,42) {3};
\node[sub, draw=am, minimum width=26mm] (ledger) at (123,32) {\textbf{reservation ledger}\\ per model: peak KV of\\ every admitted program};
\node[sub, draw=am, minimum width=26mm] (admit) at (123,15) {\textbf{admission}\\ dispatch to backend B\\ to reduce load imbalance};
\node[sub, draw=am, minimum width=26mm, fill=none] (res) at (123,2.5) {reserve at start,\\ release at end};

\node[lab] at (158,44) {$\mathcal{M}$: one model per GPU};
\node[box, minimum width=26mm, minimum height=13mm] (mA) at (158,34) {};
\node[anchor=north west, font=\footnotesize\bfseries] at (145.5,39.5) {model A backend};
\node[box, minimum width=26mm, minimum height=13mm] (mB) at (158,15) {};
\node[anchor=north west, font=\footnotesize\bfseries] at (145.5,20.5) {model B backend};
\foreach \y/\f in {34/0.82, 15/0.35}{
  \node[lab, anchor=west] at (146,\y-2) {KV};
  \draw[rule, fill=white] (151,\y-3.2) rectangle (169,\y-0.8);
  \draw[am, fill=aml] (151,\y-3.2) rectangle (151+18*\f,\y-0.8);
}

\node[box, minimum width=66mm, minimum height=12mm] (ver) at (139,-9) {};
\node[anchor=north west, font=\footnotesize\bfseries] at (107.5,-3.5) {Verifiers $\mathcal{V}$ \textcolor{mut}{\normalfont(off the critical path)}};
\node[sub] (v1) at (117,-11) {patch sandbox};
\node[sub] (v2) at (136,-11) {state checker};
\node[sub] (v3) at (156,-11) {container tests};
\node[lab] at (168.5,-11) {$\cdots$};

\draw[req] (harness.east) -- node[lab, above]{request} (new.west);
\draw[req] (new.north) -- node[lab, right]{no} (table.south);
\draw[req] (new.east) -- (emb.west); \node[lab, anchor=south] at (56.7,23.6) {yes};
\draw[req] (emb.east) -- (67.5,21) -- (67.5,31) -- (vsel.west);   %
\draw[req] (vsel.south) -- node[lab, right]{$v_{ID}$} (bandit.north);
\draw[req] (bandit.east) -- (103,11) -- (103,15) -- (admit.west);
\draw[req, rb] (admit.east) -- (141.5,15) -- (141.5,34) -- (mA.west);
\draw[req, rb] (admit.east) -- (mB.west);
\draw[sig] (ledger.east) -- (139.5,32) -- (139.5,37.5) -- (145,37.5);
\draw[sig] (ledger.east) -- (139.5,32) -- (139.5,18.5) -- (145,18.5);
\draw[fb] (mA.east) -- (174,34) -- (174,-9) -- (ver.east);
\draw[fb] (mB.east) -- (174,15);
\node[lab, text=gn, anchor=north east, align=right] at (173,7.5) {last request returns:\\ slot released, $v_{ID}$ runs};
\draw[fb] (ver.west) -- (86,-9) -- (bandit.south);
\node[tag] at (86,-0.5) {2};
\node[lab, text=gn, anchor=north] at (84,-10) {verdict $a_{ID}$, cost $\$_{ID}$\\ $\Rightarrow r^{(\alpha)}_{ID}$ updates $m_{ID}$};
\node[
  lab,
  font=\footnotesize,
  anchor=north west,
  align=center,
  text width=170mm,
  inner sep=0pt
] at (1,-17) {
  \tikz\draw[req] (0,0)--(4mm,0); request path
  \hspace{2mm}
  \tikz\draw[sig] (0,0)--(4mm,0); scheduler signal
  \hspace{2mm}
  \tikz[baseline=(t.base)]\node[tag] (t) {1}; adaptive verification
  \hspace{2mm}
  \tikz[baseline=(t.base)]\node[tag] (t) {2}; delayed feedback
  \hspace{2mm}
  \tikz[baseline=(t.base)]\node[tag, fill=am] (t) {3}; dispatch scheduler
};
\end{tikzpicture}}
\caption{\textbf{Overview of \NAMEnospace.}
The routing abstraction
$\mathcal{R}=\langle \mathrm{ID},\mathcal{P},
\mathcal{M},\mathcal{V}\rangle$
binds each program to a model and verifier.
Task classification selects the verifier from the initial request
(\circled{1}), and subsequent requests reuse the binding stored
in the lookup table.
When execution completes, the slot is released and verification
runs asynchronously; delayed feedback updates the model-selection
policy (\circled{2}).
Before execution, DISC (\circled[am]{3}) controls admission through
per-backend KV-cache reservations and selects an alternate backend
only when the predicted reward gain from reduced admission wait
exceeds the estimated reward loss from lower accuracy.}

\vspace{-5mm}
  \label{fig:method}
\end{figure}

To address the challenges described in Sec.~\ref{sec:challenges}, we present
\NAME (Fig.~\ref{fig:method}), a concurrency-aware online routing
mechanism for heterogeneous agentic workloads.
Sec.~\ref{subsec:abstraction} formalizes the framework for routing
and scheduling. Secs.~\ref{subsec:adap_verifier}
and~\ref{subsec:delayed_feedback} describe its two primary
mechanisms: adaptive verification and delayed feedback.
Finally, Sec.~\ref{subsec:codesign_scheduler} presents DISC,
a post-route dispatch scheduler that mitigates load imbalance
across heterogeneous model backends.

\subsection{Routing Abstraction}
\label{subsec:abstraction}

We abstract agentic routing using a lookup table that binds each
program to a model and a verifier. The program serves as the common
unit of concurrency, routing, and feedback. For an agentic program
$\mathcal{P}_{\mathrm{ID}}$ with identifier $\mathrm{ID}$, we define
the routing abstraction as a tuple:
\begin{equation}
    \mathcal{R} =
    \langle \mathrm{ID}, \mathcal{P},
    \mathcal{M}, \mathcal{V} \rangle,
    \label{eq:abstraction}
\end{equation}
where $\mathcal{M}=\{1,\dots,K\}$ and
$\mathcal{V}=\{1,\dots,L\}$ index the candidate models and
verifiers, respectively.

At a program's first request, \NAME assigns its unique identifier
$\mathrm{ID}$ and records the binding
$\mathcal{P}_{\mathrm{ID}}\mapsto(m,v)$ for the selected model
$m\in\mathcal{M}$ and verifier $v\in\mathcal{V}$.
Subsequent requests from $\mathcal{P}_{\mathrm{ID}}$ use the same
model, while the assigned verifier evaluates the completed program
and supplies feedback for future routing decisions.

This abstraction decouples \NAMEnospace's feedback loop from the underlying
routing policy and verifier implementation. Tool calls, host-side
computation, retrieval, and environment interactions remain part
of program execution but do not receive separate routing-table
entries.

\subsection{Concurrency-Aware Routing Mechanism}
Rapid growth in LLM usage motivates efficient concurrent program routing:
Google reports that daily requests to Gemini~\citep{team2023gemini} grew more than sevenfold between May 2025
and May 2026~\citep{pichai2026io}. Building on this lookup-table abstraction, \NAME supports online
model selection for concurrent heterogeneous agentic programs through
two mechanisms: \textit{adaptive verification} and
\textit{delayed feedback}.

\subsubsection{Adaptive Verification}
\label{subsec:adap_verifier}

Different task types require different verification of success.
Software patches can be evaluated through test execution, whereas
tool-use tasks may require verification of the resulting environment state.
Applying the same verifier across these tasks can produce incorrect
feedback, as shown in Fig.~\hyperref[fig:motivation_algo]{2a}.
\NAME therefore identifies each program's task type from its initial
request and selects a corresponding verifier.

Let $e_{\mathrm{ID}}$ denote the embedding of the initial request
of program $\mathcal{P}_{\mathrm{ID}}$.
For each verifier $v\in\mathcal{V}$, we construct a task-type
prototype $\mu_v$ by averaging the embeddings of $k$ exemplar
requests known to require that verifier. We infer the task type
by matching the incoming request to these prototypes and select
the corresponding verifier:
\begin{equation}
v_{\mathcal{P}_{\mathrm{ID}}}
=
\arg\max_{v\in\mathcal{V}}
\operatorname{sim}\big(e_{\mathrm{ID}},\mu_v\big),
\label{eq:verifier_selection}
\end{equation}
where $\mathcal{V}=\{1,\dots,L\}$ indexes the candidate verifiers
and $\operatorname{sim}$ measures embedding similarity.
This selection rule requires labeled exemplars but no additional
model training. A new verifier can be added by constructing its
prototype from the corresponding exemplar requests.

The selected verifier is recorded in the program's lookup-table
entry and evaluates its outcome upon completion. Its verdict
provides feedback for subsequent routing decisions.
Incorrect task classification can select an unsuitable verifier
and give the learner misleading feedback. We therefore evaluate
verifier-selection accuracy on held-out tasks
(Appendix~\ref{app:selector}).

\subsubsection{Delayed Feedback}
\label{subsec:delayed_feedback}

\NAME combines concurrent program dispatch with online adaptation
through \textit{delayed feedback}. When a program finishes execution,
\NAME releases its slot for the next program and runs the assigned
verifier asynchronously. Incoming programs use the current routing
policy, while verified outcomes update the policy as they become
available.

For each completed program, \NAME computes a reward combining
its verified success score and execution cost:
\begin{equation}
r^{(\alpha)}_{\mathcal{P}_{\mathrm{ID}}}
=
(1-\alpha)a_{\mathcal{P}_{\mathrm{ID}}}
-
\alpha\widetilde{\$}_{\mathcal{P}_{\mathrm{ID}}},
\label{eq:reward}
\end{equation}
where $\alpha\in[0,1]$ controls the accuracy--cost trade-off,
$a_{\mathcal{P}_{\mathrm{ID}}}\in[0,1]$ is the score returned
by the assigned verifier, and
$\widetilde{\$}_{\mathcal{P}_{\mathrm{ID}}}
=\min\left(\$_{\mathcal{P}_{\mathrm{ID}}}/\$_0,1\right)$
is the final program cost normalized by a reference cost
$\$_0>0$ and clipped at one.

Each reward is associated with the program's original routing
context and selected model through its identifier. The shared
routing policy incorporates feedback in arrival order, allowing
programs to complete and be verified in a different order from
their dispatch. Once the update is complete, \NAME removes the
program's lookup-table entry.

\subsection{Dispatch Scheduler (DISC)}
\label{subsec:codesign_scheduler}

Model quality varies across tasks. In Qwen's reported
evaluations~\citep{qwen2026qwen36}, Qwen3.6-27B leads
Qwen3.5-397B-A17B by 6.8 percentage points (pp) on
Terminal-Bench~2.0~\citep{merrill2026terminalbench}
(59.3\% vs.\ 52.5\%) but trails it by 4.7\,pp on
Humanity's Last Exam~\citep{phan2025humanity}
(24.0\% vs.\ 28.7\%). A router that learns such preferences
can concentrate tasks on one model, filling its backend's
KV cache while another remains underutilized
(Fig.~\hyperref[fig:motivation_algo]{2c}).
Addressing this imbalance requires coordinating model selection
with backend capacity. We therefore present DISC, a post-route
dispatch scheduler that adjusts the backend
assignment to reduce load imbalance.

DISC maintains a \textit{reservation ledger} for each backend.
When a program $\mathcal{P}_{\mathrm{ID}}$ is admitted to model
$m$, the ledger reserves its expected peak KV-cache footprint
and releases the reservation when execution completes.
A program is admitted only when sufficient unreserved capacity
is available. The admission component uses these reservations
to forecast the wait $w_m$ before backend $m$ has sufficient
capacity and exposes this forecast to the router at the
program boundary.

Given the routing policy's proposed model $m_{\mathrm{ID}}$
and an alternative $m'$ with a shorter forecast wait, DISC
dispatches the program to $m'$ only when the reward gain from
reduced admission wait exceeds the estimated reward loss
from lower accuracy:
\begin{equation}
\beta\frac{w_{m_{\mathrm{ID}}}-w_{m'}}{w_0}
>
(1-\beta)\big(\hat a_{m_{\mathrm{ID}}}-\hat a_{m'}\big),
\label{eq:divert}
\end{equation}
where $\beta\in[0,1]$ controls the accuracy--wait trade-off,
$\hat a_m$ is the routing policy's current estimate of model
$m$'s accuracy, and $w_0>0$ is a reference wait used for
normalization. Eq.~\ref{eq:divert} is the
expected-reward-maximizing choice under a wait-priced
reward (Appendix~\ref{app:divert_derivation}).

The dispatch decision is made once per program, before its
first request, and the final assignment is recorded in the
lookup table. Subsequent requests use the same backend,
preserving the binding in Sec.~\ref{subsec:abstraction}
without splitting the program's prefix cache across backends.

\section{Evaluation}
\label{sec:evaluation}

In this section, we evaluate \NAME on three agentic workloads (enterprise tool calling, SWE, and terminal tasks) on GH200 and H200 GPUs. Furthermore, we conduct ablation studies (Sec.~\ref{sec:eval_ablation}) to explore hyperparameters and to test \NAME on a second model pair and other GPU deployments.

\subsection{Experimental Setup}
\label{sec:eval_setup}

\textbf{Benchmarks and workflows.}
We evaluate mini-swe-agent~\citep{yang2024sweagent} on
SWE-bench Verified (108 instances for live experiments;
100 for accuracy--cost evaluation),
$\tau^2$-bench~\citep{barres2025tau2} on 276 retail,
airline, and telecom tasks, and Terminus-2 on 89
Terminal-Bench~2.0 tasks~\citep{merrill2026terminalbench}.
SWE-bench and terminal tasks have 900\,s execution budgets.
Their respective verifiers are the SWE-bench harness,
benchmark state checker, and container tests.
For $\tau^2$-bench, the selected model serves both the agent
and user simulator.

\textbf{Models and deployments.}
We use Qwen3.6-27B (FP8)~\citep{qwen3.6-27b} and
Qwen3.5-9B~\citep{qwen3.5}, one model per GPU across
two GH200 nodes. Additional experiments use a two-H200
node and a second pair, Devstral-Small-2-24B~\citep{rastogi2025devstral}
and Ministral-3-14B~\citep{liu2026ministral}.
We price input/output tokens at \$0.32/\$3.20 per million
for the 27B model and \$0.10/\$0.15 for the 9B model,
using DeepInfra reference rates~\citep{deepinfra_pricing}.
Costs include agent and user-simulator calls but exclude
classification, verification, and orchestrator routing calls.

\textbf{Model selectors.}
We evaluate LinUCB~\citep{li2010contextual} and ACRouter's
orchestrator~\citep{zhou2026acrouter}.
LinUCB uses exploration coefficient $\alpha_{\mathrm{UCB}}=0.6$
and a context combining benchmark/domain indicators from
metadata with 16 principal components of the bge-m3
initial-request embedding.
ACRouter uses Qwen3.5-4B with the released prompts and
outcome memory, evaluated at two cost settings over eight
passes and eight seeds. Memory starts empty per seed and
persists across passes, allowing retrieval of earlier
outcomes for repeated tasks
(Appendix~\ref{app:acrouter_policy}).

\begin{figure}[htbp]
\centering
\begin{minipage}[t]{0.48\textwidth}
\vspace{0pt}
\centering
\includegraphics[width=\linewidth]{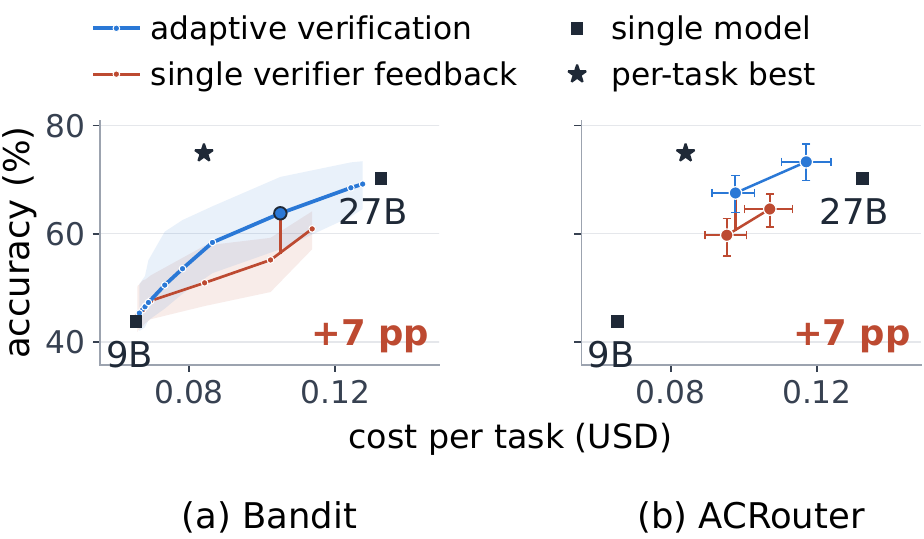}
\caption{Adaptive verification versus single-verifier feedback on (a) bandit and (b) ACRouter.}
\label{fig:eval_adap_verifier}
\end{minipage}
\hfill
\begin{minipage}[t]{0.48\textwidth}
\vspace{0pt}
\centering
\includegraphics[width=\linewidth]{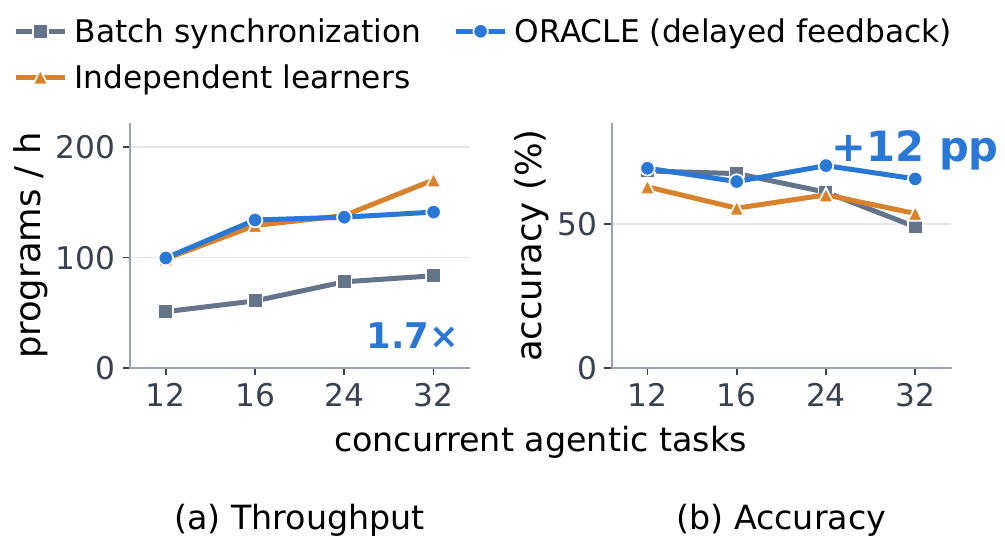}
\caption{Comparison of feedback handling on 108 SWE-bench Verified
tasks at 12--32 concurrent tasks using two GH200 GPUs.}
\label{fig:eval_delayed_feedback}
\end{minipage}
\vspace{-3mm}
\end{figure}

\textbf{Verification configurations.}
Adaptive verification uses $k=5$ exemplars per verifier.
Single-verifier feedback applies the released patch verifier
to all tasks, assigning failure feedback to $\tau^2$-bench
and Terminal-Bench tasks. Within each selector, only
verification feedback changes; evaluation always uses
benchmark-specific success signals.

\textbf{Concurrency configurations.}
We compare batch synchronization, independent per-slot
learners, and delayed feedback using the same LinUCB
algorithm and serving stack.

\textbf{Dispatch scheduling.}
DISC reserves expected peak KV-cache footprints within
85\% of the KV pool, with a 95\% per-call ceiling,
$\beta=0.5$, and $w_0=600$\,s.
Baselines include vLLM with prefix caching,
ThunderAgent's released proxy~\citep{kang2026thunderagent}
($\Delta t=5$, $f(t)=2^{-t}$), and a static cap.
The static cap admits at most 27 programs on the 27B
backend and 42 on the 9B backend, queuing additional
programs. These limits are rounded from DISC's mean
admitted counts during admission waiting at 64 concurrent tasks and remain fixed thereafter.
The static cap uses no footprint model or backend reassignment.

\textbf{Matched-cost comparisons.}
We interpolate single-verifier accuracy at the adaptive
configuration's cost and report paired task-bootstrap
confidence intervals. For the bandit, we select the
adaptive point nearest the midpoint between single-model costs.

\subsection{Adaptive Verification Evaluation}
\label{sec:eval_verifier}

\textbf{Accuracy--cost trade-off.}
Fig.~\ref{fig:eval_adap_verifier} compares adaptive verification
with single-verifier feedback using a bandit selector (a)
and ACRouter's agent selector (b). Both route between Qwen3.5-9B and
Qwen3.6-27B (FP8). The single-model baselines achieve
44\% accuracy at \$0.066 per task and 70\% at \$0.133,
respectively. Adaptive verification improves accuracy by
approximately 7 percentage points (pp) at the highlighted
matched-cost comparisons in both panels.

\textbf{Effect of verifier selection.}
Within each panel, both configurations use the same learner
and task stream; only verification feedback changes.
Single-verifier feedback applies the same verifier to all
tasks, whereas adaptive verification selects a verifier
based on task type. The task classifier selects the appropriate
verifier on 97--99\% of held-out tasks
(Appendix~\ref{app:selector}).
The gains across both routing implementations show that
task-appropriate verification improves model selection
without replacing the underlying routing algorithm.

\begin{figure}[htbp]
    \centering
    \includegraphics[width=\textwidth]{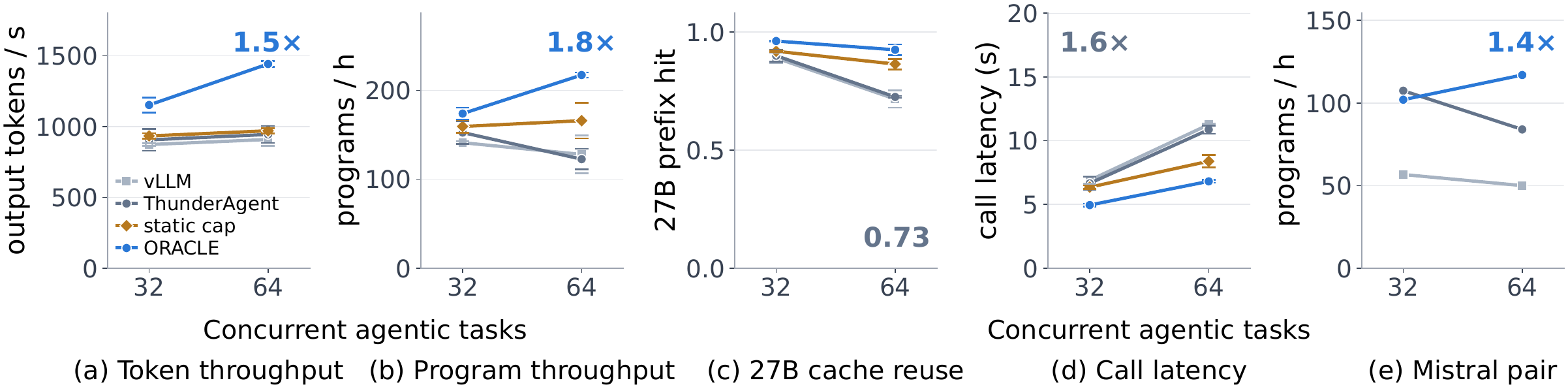}
\caption{DISC evaluation on 108 SWE-bench Verified
programs using two GH200 GPUs.
Panels (a)--(d) average two plan seeds for the Qwen pair;
panel (e) uses one seed for the Mistral pair.}

\vspace{-5mm}
    \label{fig:eval_codesign}
\end{figure}

\subsection{Delayed Feedback Evaluation}
\label{sec:eval_delayed}

\textbf{Higher throughput without batch synchronization.}
Fig.~\ref{fig:eval_delayed_feedback} evaluates one pass over
108 SWE-bench Verified tasks on the GH200 pair at 12--32
concurrent tasks. All configurations use the same routing
algorithm and serving stack, differing in how they handle
feedback. By releasing each slot when program execution
finishes, \NAME allows subsequent tasks to proceed without
waiting for other tasks or verification. At 32 concurrent tasks, it
achieves $1.7\times$ the throughput of batch-synchronous
dispatch. Its throughput is comparable to independent
learners at 12--24 concurrent tasks, but lower at 32.

\textbf{Higher accuracy through shared feedback.}
Independent per-slot learners each receive only a fraction
of the completed-task feedback, whereas \NAME uses every
verified outcome to update a shared routing policy.
Across concurrency levels, \NAME resolves 7--13 additional
tasks per 108-task run compared with independent learners.
Relative to batch synchronization, it resolves 10 and 18
additional tasks at 24 and 32 concurrent tasks, respectively.
Batch synchronization delays policy updates until the
entire batch completes, reducing opportunities to adapt
during a fixed task stream. Delayed feedback instead
incorporates outcomes as they become available, improving
accuracy while sustaining concurrent dispatch.

\subsection{Dispatch Scheduler Evaluation}
\label{sec:eval_codesign}

\textbf{Higher throughput under concurrency.}
Fig.~\ref{fig:eval_codesign} evaluates 108 SWE-bench Verified
programs on two GH200 GPUs at 32 and 64 concurrent tasks.
Panels (a)--(d) show the Qwen pair, averaged over two plan seeds.
At 64 concurrent tasks, \NAME achieves $1.8\times$ the program
throughput and $1.5\times$ the output-token throughput of
ThunderAgent. Its 27B prefix-cache hit rate remains approximately
0.93, compared with 0.73 for ThunderAgent, 0.72 for vLLM,
and 0.87 for the static cap. Mean call latency falls to 6.8\,s,
compared with 10.9\,s, 11.3\,s, and 8.4\,s, respectively.

\textbf{KV-aware admission across model pairs.}
DISC reserves each program's expected peak KV-cache footprint
and can assign it to an alternate backend before execution.
In the tested Qwen configuration, ThunderAgent's smaller
per-call KV estimate admits nearly as many calls as vLLM,
providing little protection against cache contention.
Its admission control is more effective on the Mistral pair:
the approximately 44K-token per-call estimate limits the
Devstral-24B backend to 24 simultaneous calls at 32 concurrent
tasks, compared with 31 under vLLM. Preemptions decrease from
21 to 7, and prefix-cache hit rate rises from 0.54 to 0.87.
DISC's program-level reservations maintain a hit rate of
0.99 at both concurrency levels. At 64 concurrent tasks, \NAME achieves
117 programs/h, compared with 84 for ThunderAgent and 50
for vLLM, yielding $1.4\times$ and $2.3\times$ higher
throughput, respectively (panel (e), one plan seed).

\textbf{Admission overhead.}
Reservations introduce waiting at lower concurrency.
At 32 concurrent tasks on the Qwen pair, \NAME incurs
approximately two minutes of mean admission wait.
On the Mistral pair at 32 concurrent tasks, its throughput is slightly
below ThunderAgent's (102 vs.\ 107 programs/h), although
it leads at 64 concurrent tasks. The benefits of reservation-based
admission are therefore most pronounced under higher
cache contention.

\begin{figure}[htbp]
    \centering
    \includegraphics[width=0.9\linewidth]{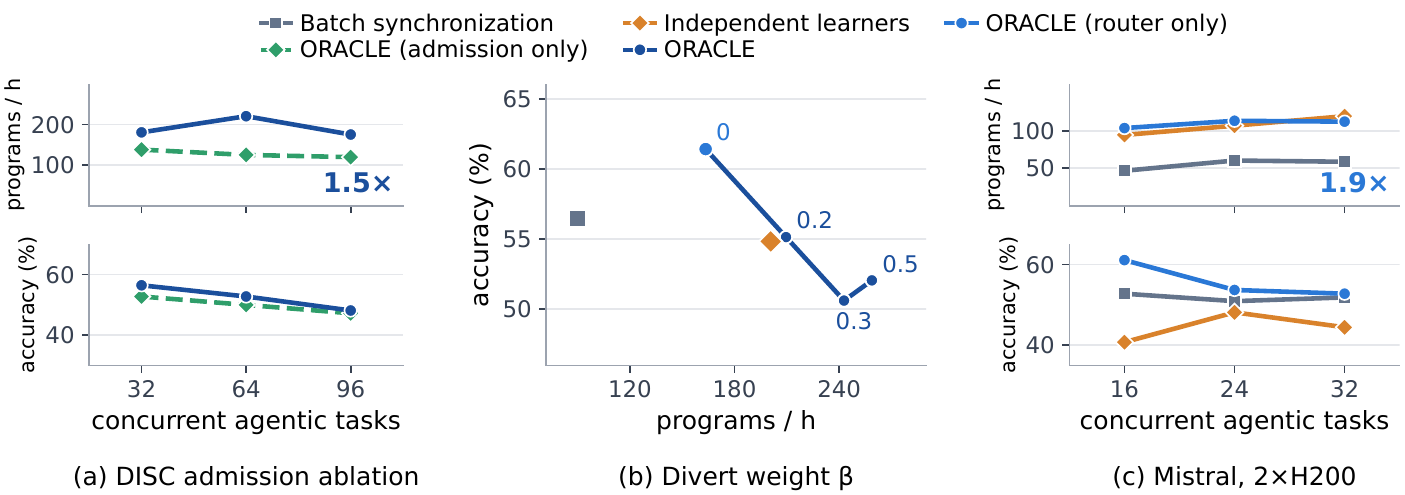}
    \caption{Ablations on SWE-bench Verified.
    Panels (a)--(b) use the Qwen pair on two GH200 GPUs;
    (b) aggregates 972 task executions.
    Panel (c) uses the Mistral pair on two H200 GPUs.}
    \label{fig:ablation}
    \vspace{-4mm}
\end{figure}

\subsection{Ablation Study}
\label{sec:eval_ablation}

\begin{wraptable}{r}{0.52\textwidth}
\vspace{-2\baselineskip}
\centering
\footnotesize
\setlength{\tabcolsep}{2pt}
\caption{Task-success gain over equal-probability model selection
at $\alpha=0$ (pp; 95\% task-bootstrap intervals).}
\label{tab:eval_verifier}
\begin{tabular}{lrrr}
\toprule
Task type & Patch-only & Adaptive & Best fixed \\
\midrule
SWE-bench & +17.9 & +17.9 [+13.0, +22.2] & +18.0 \\
$\tau^2$-bench & $-0.2$ & +8.9 [+4.1, +16.4] & +9.4 \\
TB\,2.0 & +0.1 & +9.9 [+4.2, +13.8] & +10.0 \\
\bottomrule
\end{tabular}
\end{wraptable}

\textbf{Verification across task types.}
Table~\ref{tab:eval_verifier} reports routing gains by task
type. Patch-only feedback yields almost no improvement
on $\tau^2$-bench or Terminal-Bench~2.0, whereas adaptive
verification yields gains of 8.9 and 9.9\,pp, respectively.
It also matches the 17.9\,pp gain on SWE-bench.
Across all three task types, adaptive verification is
within 0.5\,pp of the best fixed verifier chosen separately
for each type. Additional Terminal-Bench~1 results appear
in Appendix~\ref{app:verifier}. \textbf{Contribution of backend reassignment.}
Fig.~\ref{fig:ablation}(a) disables alternate-backend dispatch
while retaining DISC's reservation-based admission.
Admission alone resolves within four tasks of full \NAME
at every concurrency level, including 51 versus 52 tasks
at 96 concurrent tasks. However, full \NAME achieves
$1.5\times$ its throughput at that concurrency.
Admission control therefore preserves comparable accuracy,
while backend reassignment increases throughput by using
spare capacity rather than waiting at the preferred backend. \textbf{Dispatch-weight sensitivity.}
Fig.~\ref{fig:ablation}(b) varies $\beta$ at 64 concurrent
tasks over three stream seeds. At $\beta=0$, \NAME achieves
the highest accuracy but only $0.83\times$ the throughput
of independent learners. Setting $\beta=0.2$ gives comparable
accuracy and throughput to independent learners.
Values of 0.3 and 0.5 increase throughput to
$1.2$--$1.3\times$ their rate, at the cost of 27--41 fewer
resolved tasks across 972 executions. We use $\beta=0.5$
by default. Sensitivity to the LinUCB exploration coefficient
$\alpha_{\mathrm{UCB}}$ is reported in Appendix~\ref{app:alpha}. \textbf{Additional models and hardware.}
Fig.~\ref{fig:ablation}(c) evaluates delayed feedback with
Devstral-Small-2-24B and Ministral-3-14B on two H200 GPUs.
At 24 and 32 concurrent tasks, \NAMEnospace~(router only) resolves
6 and 9 additional tasks compared with independent learners.
At 32 concurrent tasks, it achieves $1.9\times$ the throughput of
batch synchronization, although its throughput is slightly
below that of independent learners. Additional DISC
evaluations on the Mistral pair, the Qwen pair on H200 GPUs,
and both models sharing one GH200 appear in
Appendix~\ref{app:pools}.

\section{Conclusion}
We introduced \NAMEnospace, a concurrency-aware routing framework for
agentic programs. Its lookup-table abstraction binds each program to a
model and verifier, providing a common unit for routing, feedback, and
scheduling. Adaptive verification selects task-appropriate feedback,
delayed feedback removes verification from the dispatch critical path,
and DISC uses per-backend KV-cache reservations to balance accuracy
against admission delay. Across SWE-bench, $\tau^2$-bench, and
Terminal-Bench~2.0, \NAME improves the accuracy--cost frontier by up to
7\,pp. With DISC, \NAME improves program throughput by up to
$1.8\times$ over ThunderAgent.

\subsubsection*{Acknowledgments}
We would like to thank Ishita Vohra for her contributions to technical discussions and experiments. This work was supported in part by the IBM PhD Fellowship and CoCoSys, one of seven centers in JUMP 2.0, a Semiconductor Research Corporation (SRC) program sponsored by DARPA.

\subsection*{AI use statement}
In this work, we used generative AI tools for designing and giving feedback on experiments (pre-registering the experimental blocks, their readings, stop rules and sweeps), for implementing methods (the replay and figure scripts), for refining hypotheses about the mechanisms behind the measured effects, for interpreting results, and for assisting in the writing of the derivation in Appendix~\ref{app:divert_derivation}, which formalizes a rule the authors specified. We have not used generative AI tools to generate synthetic datasets, to clean or reformat datasets, or for translation, and qualitative or thematic data analysis is not applicable to this work. Additionally, we used generative AI tools to draft the experimental and appendix sections from the measurement logs, to identify and summarize related work, to source the public API price lists behind the cost axis, to create the figures from the logged results, and to edit for readability the sections whose first drafts the authors wrote. We have reviewed all AI-assisted work: every number in the paper is read from result files by scripts the authors ran; the code used in the experiments was exercised against pre-registered readings before any reported cell ran; the derivation in Appendix~\ref{app:divert_derivation} was checked line by line by the authors; and AI-identified related work and prices were verified against the cited papers and providers' pages by the authors. We take responsibility for the final content of this work, including text, claims or artifacts produced with the aid of generative AI.

\subsection*{Reproducibility statement}
All benchmarks (SWE-bench Verified, $\tau^2$-bench, Terminal-Bench~2.0) and all model weights (Qwen3.6-27B-FP8, Qwen3.5-9B, Devstral-Small-2-24B, Ministral-3-14B) are publicly available; Section~\ref{sec:eval_setup} names the harnesses, the price table, the pools and the baselines, and Appendix~\ref{app:impl} gives the serving stack with version numbers, the single header an agent harness needs to adopt \NAMEnospace, the ledger's calibration priors and learned allowances, the verifier selector and its exemplar protocol (Appendix~\ref{app:selector}), the replay protocol with its context features, reward and bootstrap, and the grading pipeline for every benchmark. Every constant of \NAME and of the baselines was fixed before the first cell ran and is listed in Table~\ref{tab:constants}; each experimental block was pre-registered in a dated file with its cells, its reading and its stop rule, and blocks whose reading was not met are reported (Appendices~\ref{app:cold_start}). The seeds, task counts and random orders behind every figure are stated in its caption, in Section~\ref{sec:eval_setup} or in Appendix~\ref{app:impl}, and the paired bootstrap that produces every interval is described in Appendix~\ref{app:impl}. The hypotheses and the complete proof behind the divert rule are in Appendix~\ref{app:divert_derivation}.

\newpage

\bibliography{iclr2027_conference}
\bibliographystyle{iclr2027_conference}

\newpage
\appendix
\section{Extended Study}
\subsection{Request-level routing}

Existing LLM routers primarily make routing decisions at the request level. Cascade-based methods initially use cheaper backends and invoke a more capable one when confidence is low \citep{chen2024frugalgpt,aggarwal2024automix}, while learned routers predict the appropriate backend from the prompt, difficulty labels, preference data, or backend-performance models \citep{ding2024hybrid,ong2025routellm,sakota2024flyswat}, with LLMRouterBench providing a common evaluation framework \citep{li2026llmrouterbench}. Other approaches use contextual bandits that update routing decisions from per-request feedback \citep{wang2025mixllm} or jointly optimize backend selection and routing \citep{li2026stageroute}. Despite these differences, these methods treat each request as an independent routing unit and observe its reward when the request completes. This assumption does not hold for agentic tasks, where a single task may generate tens to hundreds of requests and its final outcome is known only after the entire task completes. Consequently, request-level routers do not receive direct task-level feedback to learn from earlier routing decisions.

\subsection{Bandits with delayed feedback}

\NAME uses a contextual UCB policy \citep{lattimore2020bandit}, where the reward for a routing decision becomes available only when the corresponding agentic task completes. Since the feedback delay is determined by task execution time, \NAME operates in the delayed-feedback bandit setting, where prior work shows that stochastic bandits can tolerate delayed rewards with an additive cost \citep{joulani2013delayed}. We make no new regret claim; instead, this result motivates a single learner that can continue making routing decisions while previous tasks are still running, enabling learning across concurrent tasks. In contrast, ACRouter's alternatives, such as closed-loop batching or maintaining a separate learner for each concurrency slot, either reduce throughput or limit learning across tasks.

\subsection{Multi-Model Serving}

Multi-model serving on a shared GPU pool requires both model placement and memory management. Prior systems address these challenges through model multiplexing, KV-cache and weight sharing, request migration, token-level switching, and dynamic memory allocation \citep{li2023alpaserve,duan2024muxserve,sun2024llumnix,xiang2025aegaeon,yu2025prism}, while recent work also considers KV-cache-based admission control for long-lived agentic sessions \citep{chen2026concur,cheng2026kvreservation}. These approaches generally treat the routing decision as an input and focus on managing the workload after requests are assigned to a backend. In contrast, \NAME incorporates backend memory pressure directly into routing: its reservation ledger tracks KV-cache occupancy, estimates the resulting waiting time at each backend, and uses this forecast together with model accuracy to select a backend.

\newpage
\section{Implementation Details}
\label{app:impl}

\subsection{Serving stack and interfaces}
Each model runs in its own vLLM engine behind a lightweight HTTP forwarder that the agents call in place of the engine. The forwarder logs every call's prompt, cached and completion tokens, and passes the program's key, an \texttt{X-Session-ID} header that the agent harness sets once per program, to the scheduler, so ThunderAgent's proxy and DISC's ledger see one stable key per program. Adopting \NAME therefore changes one header in the agent harness and nothing in the agent or the engine. The router runs in the experiment driver: it binds a program to a model at the program boundary, hands the program to that model's admission queue, and receives the verdict from the grader when it lands. ThunderAgent's released proxy runs in front of each engine with $\Delta t=5$\,s and priority decay on, the configuration of its paper; DISC adds the reservation ledger to the same proxy. ThunderAgent's result does not depend on the decay setting: with decay off, at 32 concurrent SWE-bench tasks on two plan seeds on the calibration pool (27B at 8\,GiB of KV), the 27B's hit rate is 0.010 and 0.023 against 0.006 and 0.021 with decay on, call p95 106 and 77\,s against 80 and 85\,s, patches 9 and 8 against 10 and 6, preemptions 54 and 47 against 67 and 47.

\subsection{Reservation ledger and divert rule}
The ledger of engine $m$ keeps, per admitted program $p$, a reservation of $\max(\text{tokens}_p, \hat c_m)+\hat\delta_m$ tokens, where $\hat c_m$ is the running mean of finished programs' peak context, started from the calibration prior $c_0$ with five pseudo-observations ($\hat c_m = (5c_0 + \sum \text{peaks})/(5+n)$; $c_0 = 22.6$k tokens on the 27B and 44.7k on the 9B), and $\hat\delta_m$ is a learned per-request allowance: on the hybrid-attention engines each in-flight request holds a fixed KV state regardless of its token count, measured offline at 3.6--4.7k token-equivalents per running request (usage $=$ tokens/capacity $+$ this constant, $R^2$ 0.79--0.91), which a token-sum capacity model such as ThunderAgent's does not see. The allowance is identifiable only while the engine is not saturated: added to ThunderAgent's proxy without the ledger, the estimator clamps to zero under the saturated pool and the arm matches ThunderAgent (27B hit rate 0.02--0.10 at 32 concurrent tasks on two plan seeds), whereas under DISC's admission control it converges to 3.0--3.4k tokens; it is a component of DISC, not a standalone fix. A new program is admitted, strictly FIFO, when the reserved total plus its own reservation fits within 85\% of the engine's KV capacity; a 95\% ceiling per call remains as the hard limit. The forecast wait is $w_m=0$ if the queue is empty and the program fits now, otherwise $w_m=(q_m+1)\,D_m/\max(K_m,1)$, with $q_m$ programs queued ahead, $K_m$ admitted and $D_m$ the mean run time of finished programs (prior 642\,s and 639\,s from the calibration passes). The divert rule of Eq.~\ref{eq:divert}, derived in Appendix~\ref{app:divert_derivation}, is evaluated once, at the boundary, with $\beta=0.5$ and $w_0=600$\,s; a diverted program starts on the other model from scratch, no KV state moves, and no program is moved after it starts. The reservation is released when the program's last call returns.

\subsection{Verifier selector}
\label{app:selector}
Every task's admission context is embedded once by bge-m3 (sentence-transformers CLS pooling, normalized, on CPU). A verifier class is represented by the mean embedding of $k=5$ exemplars drawn from tasks of that class and removed from the evaluation set; the selector returns the class of the nearest prototype by cosine. Terminal-Bench and Terminal-Bench~2.0 share the container-tests class. Over 200 random exemplar draws the per-family accuracy is 98.0\% [95, 100] on SWE-bench, 99.0\% [87, 100] on $\tau^2$-bench, 93.6\% [85, 99] on Terminal-Bench and 97.4\% [92, 100] on Terminal-Bench~2.0 ($n$ = 100, 200, 80, 89). With $k=1$ the two terminal families fall to 76\%; with $k=10$ every family is at or above 96\%. In the replays a task meets the wrong verifier with probability $1-q$ for its family's measured $q$ and is then credited 0.

\subsection{Replay protocol}
The frontier and verifier figures replay the router over measured per-task outcomes and costs of both models: $\tau^2$-bench from single-model runs of all 276 tasks, SWE-bench Verified from single-model runs over 100 instances, Terminal-Bench~2.0 from four runs per model over 89 tasks (success is the resolved rate over runs, cost the mean tokens per run at the price table). The learner is LinUCB with width $\alpha_{\mathrm{UCB}}=0.6$; its context is a bias, a family one-hot and a $\tau^2$ domain one-hot, plus 16 principal components of the task's bge-m3 embedding where one exists (the 89 Terminal-Bench~2.0 tasks and 23 SWE instances). The reward is $r=\text{success}-a\cdot\text{cost}/\text{scale}$, with scale the stream's mean 27B cost per task and $a\in\{0,0.1,0.25,0.5,0.75,1,1.5,2,3\}$. This is equivalent to the unclipped form of the reward in Eq.~\ref{eq:reward} with $\alpha=a/(1+a)$ and $\$_0=\text{scale}$, divided by the positive constant $1-\alpha$, which leaves the ordering of the two models unchanged; the live loops learn at $\alpha=0$, where the clip is inactive. Each point is the mean over 100 random stream orders; the steady state (Fig.~\ref{fig:eval_adap_verifier}) runs eight passes over the stream and scores the last, the cold start (Appendix~\ref{app:cold_start}) one pass. Intervals are task bootstraps stratified by family, 200 draws. Random mixing is the line between the two single models; the per-task best picks the cheaper successful model for each task. Fig.~\ref{fig:eval_adap_verifier}(a) and Fig.~\ref{fig:frontier_tau2_swe}(a) use this replay with LinUCB; panel (b) of each replaces LinUCB with ACRouter's orchestrator (Appendix~\ref{app:acrouter_policy}) and keeps everything else.

\subsection{Harnesses and grading}
SWE-bench programs run mini-SWE-agent 2.2.8, one container per program with a 900\,s budget, on the node that hosts the 9B (arm64 instance images under Docker) or, on 2xH200, under rootless Podman with x86 images. Verdicts come from the SWE-bench harness (swebench 5.0.2), run after the cell for the scheduling experiments and inline by a thread pool for the loop experiments, where the verdict's arrival time is the delayed feedback under study; on 2xH200 a gold-patch guard grades two known instances before any cell is trusted. $\tau^2$-bench runs the benchmark's own simulation with its user simulator and judge served by the local engines, the routed model playing the user of its own conversation, and its state checker as the verdict. Terminal-Bench~2.0 runs Terminus-2 through harbor 0.22.0, one harbor run per task, 24 concurrent tasks, 900\,s per task. Program throughput counts the programs that submit their result (a patch, for SWE-bench) inside their budget, per hour of the cell's makespan. Accuracy counts a task as resolved only if its program finished inside its budget and the verifier accepted it.

\subsection{Pre-registration}
Every block, the frontier, the loop sweep, the scheduler comparison and each ablation, was written up in a dated file before its first cell ran, with the cells, the reading that would count as met, and a stop rule. Every constant in Table~\ref{tab:constants} was fixed before the first cell and never changed after. Blocks whose reading was not met are reported as such, for example the cold-start frontier (Appendix~\ref{app:cold_start}).

\begin{table}[htbp]
\centering
\small
\caption{Constants, all fixed before the first cell.}
\label{tab:constants}
\begin{tabular}{llr}
\toprule
Component & Constant & Value \\
\midrule
router & LinUCB width $\alpha_{\mathrm{UCB}}$ & 0.6 \\
router & tie & cheaper model \\
verifier selector & exemplars per class $k$ & 5 \\
ACRouter orchestrator & memory neighbors / similarity floor & 10 / 0.5 \\
ACRouter orchestrator & cost weights & $-0.1$, $-50$ \\
ledger & admission fraction / per-call ceiling & 0.85 / 0.95 \\
ledger & prior pseudo-observations & 5 \\
ledger & peak-context prior $c_0$ (27B / 9B) & 22.6k / 44.7k tokens \\
ledger & run-time prior $D_0$ (27B / 9B) & 642 / 639\,s \\
divert rule & $\beta$, $w_0$ & 0.5, 600\,s \\
ThunderAgent proxy & $\Delta t$, decay & 5\,s, $2^{-t}$ \\
static cap & admitted programs (27B / 9B) & 27 / 42 \\
agents & episode budget & 900\,s \\
\bottomrule
\end{tabular}
\end{table}

\newpage

\section{Extended Theoretical Analysis}
\subsection{Derivation of the Divert Rule}
\label{app:divert_derivation}

This appendix shows that the divert rule in Eq.~4 is the choice with the highest expected reward when the admission wait is priced in the same way that Eq.~3 prices cost. We first state three assumptions and check each one against our measurements. We then give the proof and two corollaries.

\begin{assumption}[Wait-priced reward]
\label{as:reward}
Let $w$ be the admission wait of a program: the time from its arrival at the task boundary until its first request is served. We divide the wait by a reference wait $w_0 > 0$ and clip the result at one,
\begin{equation}
\tilde{w} = \min\left(w / w_0,\ 1\right),
\end{equation}
and define the wait-priced reward as
\begin{equation}
r_{\mathrm{wait}} = (1 - \beta)\, a - \beta\, \tilde{w}, \qquad \beta \in [0, 1],
\end{equation}
where $a \in [0, 1]$ is the graded accuracy of the program. The clip keeps the penalty bounded for very long waits, as the clip on cost in Eq.~3 does. It has no effect while $w < w_0$.
\end{assumption}

\begin{assumption}[Forecast waits below the reference]
\label{as:waits}
At the task boundary, the reservation ledger of Appendix~B.2 gives a forecast wait $w_m$ for each candidate model $m$, and both forecasts are below the reference wait, $w_m < w_0$. In our runs $w_0 = 600$~s and the mean admission waits in Table~3 are between 86 and 130~s, so the assumption holds on average. Corollary~\ref{cor:saturated} covers the case $w_m \ge w_0$.
\end{assumption}

\begin{assumption}[Selection estimate]
\label{as:estimate}
For each model $m$, the router keeps an estimate $\hat{a}_m$ of the accuracy it expects from $m$ on the current program. We use the bandit's selection index as $\hat{a}_m$. Because the bandit chose $m_{\mathrm{ID}}$ by this index, $\hat{a}_{m_{\mathrm{ID}}} \ge \hat{a}_{m'}$ for every other model $m'$.
\end{assumption}

\begin{remark}[Run time]
The reward in Assumption~\ref{as:reward} prices the wait only. It does not price the run time of the program. Run time depends on the model and, under load, on the state of the engine (Section~5.4), and we do not model it here. The result below is therefore a statement about the wait-priced reward, not about end-to-end latency.
\end{remark}

\begin{proposition}[Divert rule]
\label{prop:divert}
Let Assumptions~\ref{as:reward} to~\ref{as:estimate} hold. Let $m_{\mathrm{ID}}$ be the model the bandit chose, and let $m'$ be the other model, which has the shorter forecast wait. The model with the higher expected wait-priced reward is $m'$ if and only if
\begin{equation}
\label{eq:divert-app}
\beta\, \frac{w_{m_{\mathrm{ID}}} - w_{m'}}{w_0} \;>\; (1 - \beta)\left(\hat{a}_{m_{\mathrm{ID}}} - \hat{a}_{m'}\right),
\end{equation}
which is Eq.~4.
\end{proposition}

\begin{proof}
By Assumption~\ref{as:waits}, the clip is inactive for both models, so $\tilde{w}_m = w_m / w_0$. Taking the expectation of $r_{\mathrm{wait}}$ with the estimates of Assumption~\ref{as:estimate} gives, for each candidate $m$,
\begin{equation}
\mathbb{E}\left[r_{\mathrm{wait}} \mid m\right] = (1 - \beta)\, \hat{a}_m - \beta\, \frac{w_m}{w_0}.
\end{equation}
The router should choose $m'$ when $\mathbb{E}[r_{\mathrm{wait}} \mid m'] > \mathbb{E}[r_{\mathrm{wait}} \mid m_{\mathrm{ID}}]$. Writing out both sides,
\begin{equation}
(1 - \beta)\, \hat{a}_{m'} - \beta\, \frac{w_{m'}}{w_0} \;>\; (1 - \beta)\, \hat{a}_{m_{\mathrm{ID}}} - \beta\, \frac{w_{m_{\mathrm{ID}}}}{w_0}.
\end{equation}
Moving the wait terms to the left and the accuracy terms to the right gives Eq.~(\ref{eq:divert-app}). The left side is the wait the program saves, priced at $\beta / w_0$. The right side is the accuracy it gives up, priced at $1 - \beta$.
\end{proof}

\begin{corollary}[Endpoints]
\label{cor:endpoints}
At $\beta = 0$, Eq.~(\ref{eq:divert-app}) reduces to $\hat{a}_{m'} > \hat{a}_{m_{\mathrm{ID}}}$. This never holds under Assumption~\ref{as:estimate}, so the router never diverts and acts as a quality-only bandit. At $\beta = 1$, the rule reduces to $w_{m'} < w_{m_{\mathrm{ID}}}$, so the router always diverts to the shorter wait, whatever the accuracy. The sweep in Fig.~\ref{fig:ablation}(b) follows this path: $\beta = 0$ resolves the most tasks at the lowest rate, and a larger $\beta$ trades resolved tasks for speed.
\end{corollary}

\begin{corollary}[Saturated wait]
\label{cor:saturated}
If $w_{m_{\mathrm{ID}}} \ge w_0$, the wait penalty of $m_{\mathrm{ID}}$ is clipped at $\beta$, and the rule becomes
\begin{equation}
\beta \left(1 - \frac{w_{m'}}{w_0}\right) \;>\; (1 - \beta)\left(\hat{a}_{m_{\mathrm{ID}}} - \hat{a}_{m'}\right).
\end{equation}
The largest accuracy gap the rule can accept is then $\beta / (1 - \beta)$. At the deployed $\beta = 0.5$ this bound equals one, so the clip does not restrict the rule. It only keeps the penalty bounded.
\end{corollary}

\begin{remark}[Choice of $w_0$]
Below saturation, the rule depends on $\beta$ and $w_0$ only through the ratio $(1 - \beta)\, w_0 / \beta$: a wait of $w_0$ seconds trades against an accuracy gap of $\beta / (1 - \beta)$. Sweeping $\beta$ at a fixed $w_0$, as in Fig.~\ref{fig:ablation}(b), therefore also covers the choice of $w_0$.
\end{remark}

\newpage
\section{Cache Reuse and Latency Statistics}
\label{app:latency}
Table~\ref{tab:latency} lists, for plan seed 912 of Fig.~\ref{fig:eval_codesign}, the quantities behind panels (a), (c) and (d) together with the two the figure does not show: the admission wait the ledger imposes and the net change in a program's submit-to-done time against stock vLLM. Three points follow. First, output tokens per second rise under \NAME at every concurrency level, 1.2--1.7$\times$ stock vLLM, because reserved programs decode at a high prefix hit rate instead of recomputing prefixes: the 27B's hit rate is 0.96, 0.90 and 0.75 against 0.87, 0.75 and 0.28 as concurrency grows. Second, the call latency that agents see is 1.4--1.8$\times$ lower on the mean and 1.5--2.5$\times$ lower at the tail. Third, the reservation's cost is the admission wait, 1.4--2.2 minutes per program; at 32 concurrent tasks nothing is contended, so the wait is pure delay and submit-to-done grows by 5.3 minutes, whereas at 64 concurrent tasks the faster calls repay it exactly and at 96 the program finishes 37\,s sooner despite waiting 101\,s at admission.

\begin{table}[htbp]
\centering
\small
\caption{Cache reuse and latency on the GH200 pair (SWE-bench Verified, 108 tasks, plan seed 912). Call latency is the mean and p95 over every model call; wait is the mean admission wait; $\Delta$ is the mean change in submit-to-done time against stock vLLM (95\% task bootstrap).}
\label{tab:latency}
\begin{tabular}{rlrrrrrr}
\toprule
concurrent tasks & system & 27B hit & tok/s & call mean & call p95 & wait & $\Delta$ submit$\to$done \\
\midrule
32 & stock vLLM & .87 & 863 & 7.2\,s & 27.8\,s & -- & -- \\
32 & ThunderAgent & .87 & 829 & 7.2\,s & 28.9\,s & -- & $-3$\,s \\
32 & \NAME & .96 & 1204 & 5.1\,s & 18.4\,s & 130\,s & +316\,s [+255, +379] \\
\midrule
64 & stock vLLM & .75 & 864 & 11.3\,s & 43.0\,s & -- & -- \\
64 & ThunderAgent & .72 & 884 & 11.2\,s & 40.7\,s & -- & $-1$\,s \\
64 & \NAME & .90 & 1461 & 6.9\,s & 25.2\,s & 86\,s & +6\,s [$-55$, +66] \\
\midrule
96 & stock vLLM & .28 & 1049 & 15.5\,s & 72.0\,s & -- & -- \\
96 & ThunderAgent & .29 & 1063 & 15.5\,s & 64.6\,s & -- & +17\,s \\
96 & \NAME & .75 & 1237 & 8.8\,s & 28.6\,s & 101\,s & $-37$\,s [$-92$, +18] \\
\bottomrule
\end{tabular}
\end{table}

\newpage

\section{Additional Results and Ablations}
\label{app:results}

\subsection{Decomposing DISC}
\label{app:tapp}
Table~\ref{tab:tapp} lists the five systems behind Fig.~\ref{fig:ablation}(a) (GH200 pair, plan seed 912), adding the two baselines and the static per-model cap of Fig.~\ref{fig:eval_codesign}, which admits at most 27 programs on the 27B and 42 on the 9B, values fixed from plan seed 911. All three reservation-based systems keep the 27B's prefix hit rate above 0.75 at 96 concurrent tasks, where stock vLLM and ThunderAgent fall below 0.3, and all three resolve two to three times as many tasks. They differ in speed: DISC resolves 130 tasks per hour against 94 with admission only and 100 for the cap, at a similar resolved count, because a program forecast to wait on the 27B is dispatched to the idle 9B instead. The cap also needs its two values chosen per pool and workload; the ledger learns each program's peak context online.

\begin{table}[htbp]
\centering
\small
\caption{DISC decomposed on the GH200 pair, SWE-bench Verified, 108 tasks, plan seed 912.}
\label{tab:tapp}
\begin{tabular}{lrrrrrr}
\toprule
 & \multicolumn{3}{c}{64 concurrent tasks} & \multicolumn{3}{c}{96 concurrent tasks} \\
\cmidrule(lr){2-4}\cmidrule(lr){5-7}
 & resolved & res./h & 27B hit & resolved & res./h & 27B hit \\
\midrule
stock vLLM & 40 & 82 & .75 & 22 & 66 & .28 \\
ThunderAgent & 41 & 89 & .72 & 20 & 61 & .29 \\
static cap (27$\times$42) & 56 & 112 & .84 & 57 & 100 & .86 \\
\NAME (admission only) & 54 & 99 & .80 & 51 & 94 & .80 \\
\NAME & 57 & 159 & .90 & 52 & 130 & .75 \\
\bottomrule
\end{tabular}
\end{table}

\subsection{Other Pools and Model Families}
\label{app:pools}
Table~\ref{tab:pools} extends the comparison of Fig.~\ref{fig:eval_codesign} to 96 concurrent tasks, in resolved tasks per hour rather than program throughput, on a node with two H200s (141\,GB each), with the Qwen pair and with a second family, Devstral-Small-2-24B and Ministral-3-14B, and on a single GH200 hosting both models. The direction holds on every pool. The Mistral pair resolves 29 tasks per seed under \NAME against 17 under ThunderAgent and 16 under stock vLLM (+27 of 216 [+12, +42] over stock), at 1.2--1.3$\times$ their rate and with only 70\,s of added submit-to-done time. On the single GPU, where the two models share one KV cache, the gap is largest and appears at 32 concurrent tasks already: 41 resolved against 22 and 23. On the H200 pair, whose larger cache contends least, the gap is smallest, 1.3--1.5$\times$. The gain scales with the memory pressure the workload creates.

\begin{table}[htbp]
\centering
\small
\caption{\NAME against ThunderAgent and stock vLLM at 96 concurrent tasks on other pools and models: resolved tasks per hour, mean of two plan seeds (single-GPU row: 32 concurrent tasks, one seed).}
\label{tab:pools}
\begin{tabular}{lrrrr}
\toprule
 & \multicolumn{2}{c}{resolved / h} & \multicolumn{2}{c}{\NAME vs.} \\
\cmidrule(lr){2-3}\cmidrule(lr){4-5}
Pool and models & ThunderAgent & \NAME & ThunderAgent & stock \\
\midrule
2$\times$GH200, Qwen (Fig.~\ref{fig:eval_codesign}) & 60 & 127 & 2.1$\times$ & 1.9$\times$ \\
2$\times$H200, Qwen & 198 & 265 & 1.3$\times$ & 1.5$\times$ \\
2$\times$H200, Devstral-24B / Ministral-14B & 56 & 69 & 1.2$\times$ & 1.3$\times$ \\
1$\times$GH200, both models (32 concurrent tasks) & 24 & 41 & 1.7$\times$ & 1.6$\times$ \\
\bottomrule
\end{tabular}
\end{table}

\subsection{Verifier per Family}
\label{app:verifier}
\begin{figure}[htbp]
\centering
\includegraphics[width=0.95\textwidth]{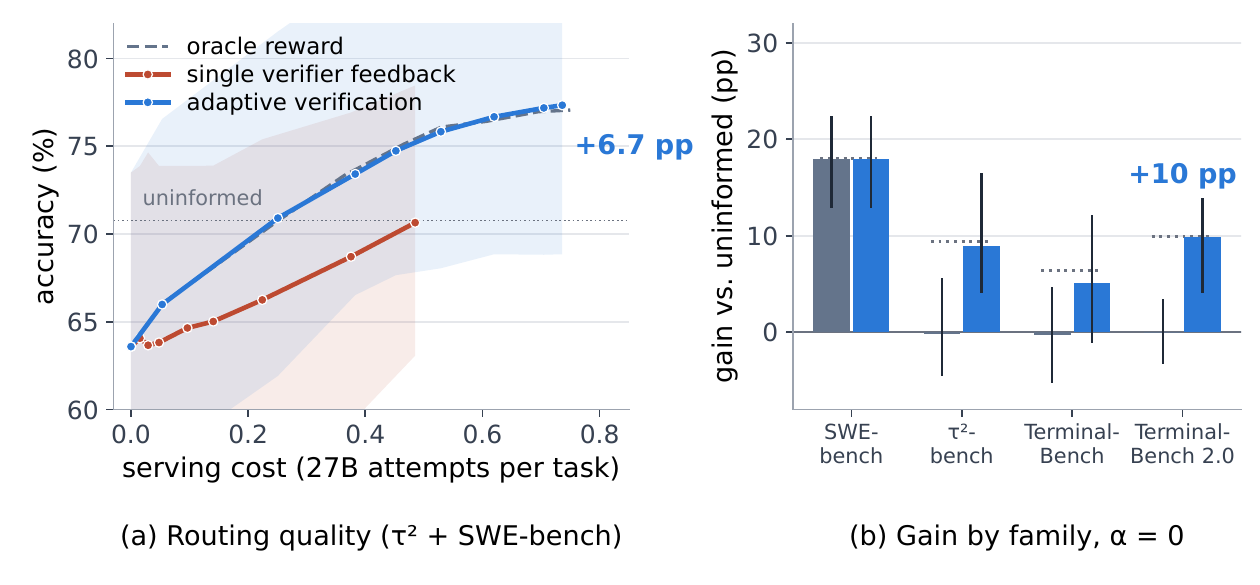}
\caption{\textbf{A single patch verifier loses every family it cannot read.} Choosing the verifier per task recovers +5 to +10\,pp on $\tau^2$-bench and both Terminal-Bench versions and loses nothing on SWE-bench. Both routed curves use the bandit selector.}
\label{fig:abl_verifier}
\end{figure}
Fig.~\ref{fig:abl_verifier}(b) is the replay behind Table~\ref{tab:eval_verifier}, each family alone at cost weight 0. With the single patch verifier the gain over uninformed routing is $-0.2$, $-0.2$ and $+0.1$\,pp on $\tau^2$-bench, Terminal-Bench and Terminal-Bench~2.0: the router receives the same verdict for every outcome and learns nothing. The adaptive verifier gains +8.9 [+4.1, +16.4], +5.1 [$-1.0$, +12.1] and +9.9 [+4.2, +13.8]\,pp, within 1.5\,pp of the best fixed verifier for each family, and matches it on SWE-bench (+17.9\,pp), the one family the patch verifier reads. Panel (a) shows the same effect on a mixed $\tau^2$+SWE stream against the share of tasks sent to the 27B: single-verifier feedback tops out at 71\% success, adaptive verification reaches 77\% and stays above it at every share, within 1\,pp of oracle reward.

\subsection{Adaptive Verification on the $\tau^2$ + SWE Stream}
\label{app:frontier_tau2_swe}
\begin{figure}[htbp]
\centering
\includegraphics[width=0.7\textwidth]{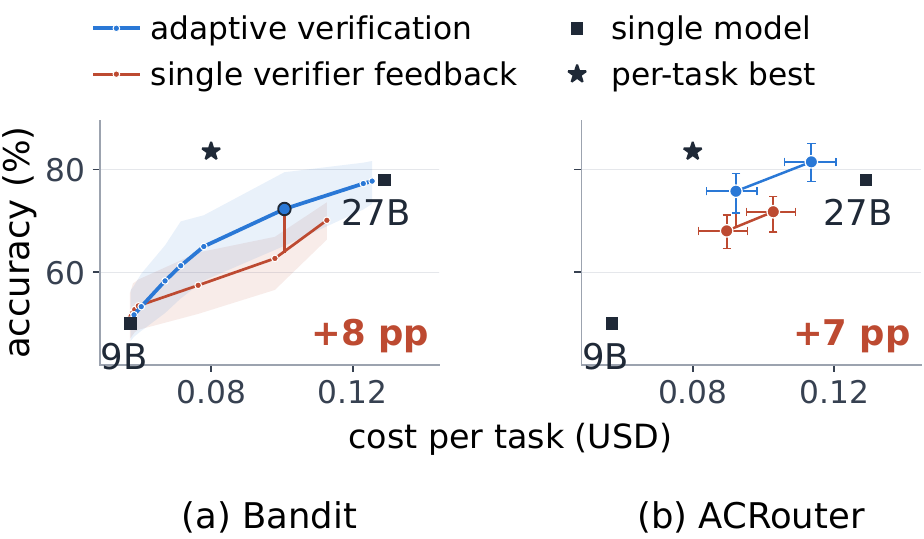}
\caption{Adaptive verification versus single-verifier feedback on the $\tau^2$+SWE stream with (a) the bandit selector and (b) ACRouter's agent selector, in the layout of Fig.~\ref{fig:eval_adap_verifier}.}
\label{fig:frontier_tau2_swe}
\end{figure}
Fig.~\ref{fig:frontier_tau2_swe} repeats Fig.~\ref{fig:eval_adap_verifier} on the stream of 276 $\tau^2$-bench and 100 SWE-bench Verified tasks, without Terminal-Bench~2.0. The single models reach 50.0\% accuracy at \$0.057 per task (9B) and 77.9\% at \$0.129 (27B); the per-task best is 83.5\%. With the bandit selector, adaptive verification is +8.3\,pp [+2.7, +15.1] above single-verifier feedback at the matched cost of \$0.101. With ACRouter's orchestrator, adaptive verification reaches 75.8\% [71.5, 79.2] at \$0.092 and 81.5\% [77.6, 85.0] at \$0.114 at the two cost settings, against 68.1\% [64.6, 71.2] at \$0.090 and 71.8\% [67.8, 74.8] at \$0.103 with single-verifier feedback, a gain of +7.0\,pp [+2.7, +10.4] at matched cost. Both gains are slightly larger than on the three-family stream of Fig.~\ref{fig:eval_adap_verifier}, where $\tau^2$-bench tasks, the ones a patch verifier cannot read, make up a smaller share of the stream.

\subsection{Frontier on the SWE + Terminal-Bench~2.0 Stream}
\label{app:frontier_swe_tb2}
\begin{figure}[htbp]
\centering
\includegraphics[width=0.50\textwidth]{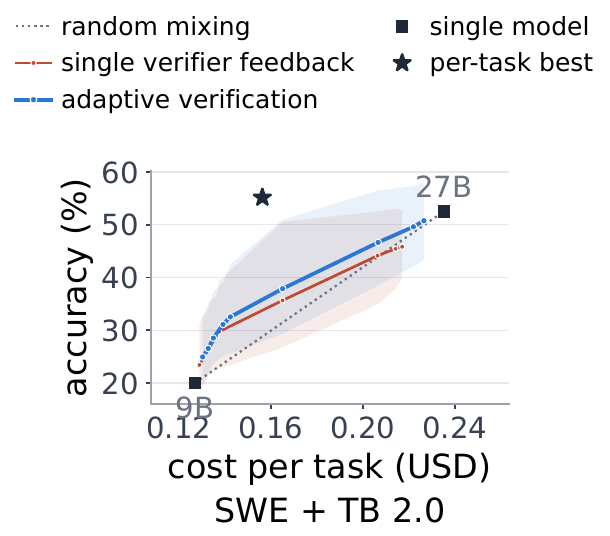}
\caption{\textbf{Accuracy--cost frontier on the SWE + Terminal-Bench~2.0 stream.} The bandit selector with adaptive verification sits +6.4\,pp above random mixing at the midpoint cost; both routed curves use the bandit selector.}
\label{fig:frontier_swe_tb2}
\end{figure}
Fig.~\ref{fig:frontier_swe_tb2} is the third stream, the two families whose verifiers run in containers; Fig.~\ref{fig:eval_adap_verifier} and Fig.~\ref{fig:frontier_tau2_swe} show the other two. The single models span 20\% at \$0.127 per task and 52\% at \$0.235; \NAME spans 25--51\% for \$0.130--\$0.227 and sits +6.4\,pp [+2.9, +12.5] above random mixing at the midpoint cost. Single-verifier feedback trails by 2--5\,pp at the same cost weight.

\subsection{ACRouter's Own Policy}
\label{app:acrouter_policy}
Fig.~\ref{fig:eval_adap_verifier}(b) and Fig.~\ref{fig:frontier_tau2_swe}(b) run ACRouter's own model selector. This section gives its setup, then two further measurements of the released system.

\textbf{Hosted orchestrator.} ACRouter's orchestrator is an LLM that chooses the model for each task from a prompt. We serve Qwen3.5-4B with vLLM on a GH200 and build the prompt as the released code does: the task's initial request, the cost-tiered model descriptions, a complexity tier from LiteLLM's rule-based classifier, and the nearest past outcomes from ACRouter's memory (up to 10 neighbors with cosine similarity of at least 0.5 between bge-m3 embeddings of the initial requests). The orchestrator answers with a JSON decision; a decision costs about 660 prompt and 7 completion tokens, and no call fell back to a default choice. Each run replays eight passes over the stream for eight seeds, one orchestrator call per task and pass. The memory starts empty for each seed and persists across passes, so from the second pass a repeated task finds its own earlier verdict among its neighbors; this recall is what lifts the orchestrator's accuracy at its released cost setting above the 27B alone. The two cost settings are the released cost weight ($-0.1$) and a cost-penalized one ($-50$): the released weight multiplies a few cents against a score in $[0,1]$ and barely moves the decision, so the second setting is needed for a second point. Adaptive verification and single-verifier feedback differ only in the verdict written to the memory, so the comparison within each panel (b) is paired.

\textbf{Released memory policy in the verifier replay.} Substituting ACRouter's released \texttt{route()} rule (per-dimension mean of past rewards, arg-max, ties to the cheap chain) for LinUCB in the replay of Fig.~\ref{fig:eval_adap_verifier}, on the $\tau^2$+SWE pool of 65 tasks it reaches 56.0\% task success with its fixed verifier and 67.0\% with the adaptive verifier, against 64.4\% and 72.7\% for LinUCB under the same two verifiers and 64.6\% for uninformed routing. On two larger pools (100 $\tau^2$ + 100 SWE tasks each) it stays within 1\,pp of uninformed routing under either verifier. The memory rule needs many verdicts per dimension before its means separate, which these streams do not supply.

\textbf{Earlier reconstruction as one operating point.} An earlier reconstruction of the complete system, an LLM orchestrator prompted with the memory's nearest neighbors and cost-tiered model descriptions, warm-started on a 30\% probing split as in its paper, lands on the per-domain accuracy--cost frontiers at one point each: 77.3\% at \$0.035 on retail, 82.7\% at \$0.082 on telecom, 82.5\% at \$0.032 on airline, 61.6\% resolved at \$0.31 on SWE-bench. Each point is on or near the frontier that \NAME traces, but it is a single point: the published cost weight is inert over its released range, which is why the hosted runs above add a cost-penalized setting. Its routing decision costs 209\,ms at the median against 0.56\,ms for \NAMEnospace's, on the critical path of every program. Two caveats: the orchestrator is a prompted 4B model rather than their fine-tuned 0.8B adapter, and the points are scored on the held-out 70\% split, which is 2.5--3\,pp more favorable than the full pools the other curves use.

\textbf{Released fine-tuned router.} Running the published 0.8B LoRA router itself, greedily, with its prompt and its eight backend names mapped to our models by its own tier labels, on 274 $\tau^2$ tasks gives $-5.1$\,pp [$-8.8$, $-1.5$] against the best single model; it uses two or three of its eight names. The published adapter does not reproduce the decisions in the paper's own traces (7\% per-task agreement on its 112 code tasks), so this bounds the release, not the paper.

\subsection{Exploration Width}
\label{app:alpha}
\begin{figure}[htbp]
\centering
\includegraphics[width=0.5\textwidth]{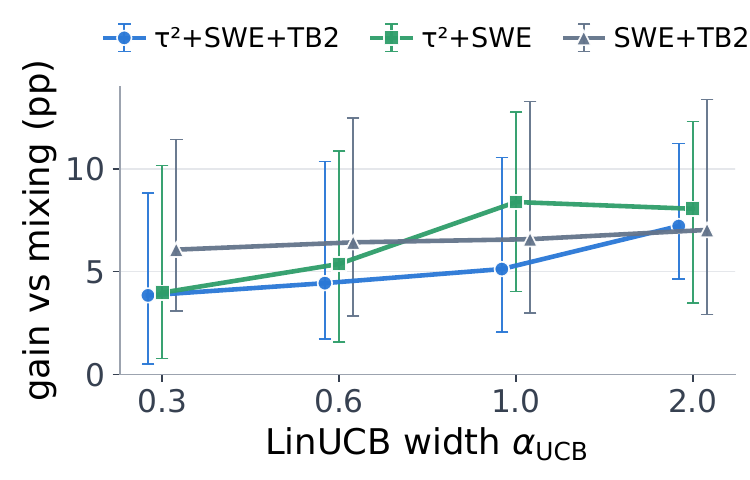}
\caption{\textbf{The gain over random mixing survives every exploration width.} Replay of Fig.~\ref{fig:eval_adap_verifier} at four LinUCB widths, \NAME minus random mixing at the midpoint cost (95\% task bootstrap); the main figures use the pre-registered $\alpha_{\mathrm{UCB}}=0.6$.}
\label{fig:alpha}
\end{figure}
Fig.~\ref{fig:alpha} sweeps the LinUCB width $\alpha_{\mathrm{UCB}}$ in the frontier replay of Fig.~\ref{fig:eval_adap_verifier}, and Table~\ref{tab:alpha} gives the intervals. Every interval excludes zero. Wider exploration helps on the two $\tau^2$-containing streams (+3.8 to +7.2\,pp on $\tau^2$+SWE+TB\,2.0, +4.0 to +8.4\,pp on $\tau^2$+SWE) and is flat on SWE + TB\,2.0 (+6.1 to +7.0\,pp). The main figures keep the pre-registered $\alpha_{\mathrm{UCB}}=0.6$.

\begin{table}[htbp]
\centering
\small
\caption{\NAME minus random mixing at the midpoint cost (pp, 95\% task bootstrap) by LinUCB width $\alpha_{\mathrm{UCB}}$.}
\label{tab:alpha}
\begin{tabular}{lcccc}
\toprule
stream & $\alpha_{\mathrm{UCB}}=0.3$ & $\alpha_{\mathrm{UCB}}=0.6$ & $\alpha_{\mathrm{UCB}}=1.0$ & $\alpha_{\mathrm{UCB}}=2.0$ \\
\midrule
$\tau^2$ + SWE + TB\,2.0 & +3.8 [+0.5, +8.8] & +4.4 [+1.7, +10.4] & +5.1 [+2.1, +10.5] & +7.2 [+4.6, +11.2] \\
$\tau^2$ + SWE & +4.0 [+0.8, +10.2] & +5.4 [+1.6, +10.9] & +8.4 [+4.0, +12.8] & +8.1 [+3.5, +12.3] \\
SWE + TB\,2.0 & +6.1 [+3.1, +11.4] & +6.4 [+2.9, +12.5] & +6.6 [+3.0, +13.3] & +7.0 [+2.9, +13.4] \\
\bottomrule
\end{tabular}
\end{table}

\subsection{Cold Start}
\label{app:cold_start}
Fig.~\ref{fig:eval_adap_verifier} reports the router's steady state after eight passes over the stream. Replayed cold, in a single pass with family and domain context only, \NAME coincides with random mixing of the two models: +0.9\,pp [$-1.4$, +4.3] at the midpoint cost on the $\tau^2$+SWE+TB\,2.0 stream, +1.6 [$-1.3$, +5.7] on $\tau^2$+SWE and $-0.8$ [$-2.1$, +2.6] on SWE+TB\,2.0. The frontier gain therefore emerges with accumulated task-level feedback; a single cold pass over the stream is insufficient to reach the steady-state improvement reported in Fig.~\ref{fig:eval_adap_verifier}.

\subsection{Pending Term in the Width}
\label{app:pending_width}
Delayed credit counts every decision made so far in the confidence width, landed or not; the alternative counts only landed verdicts. Live on the GH200 pair (108 tasks, stream seed 921, DISC stack), removing the pending term costs 7 resolved tasks at 32 concurrent tasks (64 versus 71) and 4 at 64 (54 versus 58) at the same rate, with intervals of [$-3$, +17] and [$-4$, +12] that include zero: on one seed the term is not separable. Its effect is visible in the routing rather than the totals: without it the width collapses while verdicts are in flight and the router commits earlier to the cheaper 9B (27B share 0.65 against 0.76 at 32 concurrent tasks, 0.28 against 0.53 at 64).

\end{document}